%% file: clear2026.tex
\documentclass[final,12pt]{clear2026} 

\title[IV Co-Scientist]{IV Co-Scientist: Multi-Agent LLM Framework for \\ Causal Instrumental Variable Discovery}
\usepackage{times}

\usepackage{times}
\usepackage{booktabs}     
\usepackage{multirow}     
\usepackage{xcolor}       
\usepackage{adjustbox}    
\usepackage{changepage}   

\usepackage{float}
\usepackage{xcolor} 
\usepackage{tikz}
\usepackage{bbm}
\usepackage{pifont}
\usepackage{titletoc}
\usepackage{newfloat}
\usepackage{colortbl} 
\usepackage{booktabs} 
\usepackage[skins]{tcolorbox} 
\usepackage{xcolor}         

\newtcolorbox{boxB}[1]{
  title=#1,
  fonttitle=\bfseries,
  sharp corners
}

\newtcolorbox{boxA}{
    fontupper = \bf,
    boxrule = 1.5pt,
    colframe = black 
}
\clearauthor{%
 \Name{Ivaxi Sheth} \Email{ivaxi.sheth@cispa.de}\\
 \addr CISPA Helmholtz Center for Information Security
 \AND
 \Name{Zhijing Jin} \Email{zjin.admin@cs.toronto.edu }\\
 \addr Max Planck Institute for Intelligent Systems,
Jinesis AI Lab, University of Toronto \& Vector Institute, \\ EuroSafeAI%
 \AND
 \Name{Bryan Wilder} \Email{bwilder@cmu.edu}\\
 \addr Carnegie Mellon University
 \AND
 \Name{Dominik Janzing}\thanks{Work done outside of Amazon.}\thanks{Equal supervision.} \Email{janzind@amazon.de}\\
 \addr Amazon
 \AND
 \Name{Mario Fritz}\footnotemark[2] \Email{fritz@cispa.de}\\
 \addr CISPA Helmholtz Center for Information Security
}
\begin{document}

\maketitle

\begin{abstract}%
      In the presence of confounding between an endogenous variable and the outcome, instrumental variables (IVs) are used to isolate the causal effect of the endogenous variable. Identifying valid instruments requires interdisciplinary knowledge, creativity, and contextual understanding, making it a non-trivial task. In this paper, we investigate whether large language models (LLMs) can aid in this task. We perform a two-stage evaluation framework. First, we test whether LLMs can recover well-established instruments from the literature, assessing their ability to replicate standard reasoning. Second, we evaluate whether LLMs can identify and avoid instruments that have been empirically or theoretically discredited. Building on these results, we introduce \textbf{IV Co-Scientist}, a multi-agent system that proposes, critiques, and refines IVs for a given treatment–outcome pair. We also introduce a statistical test to contextualize consistency in the absence of ground truth. Our results show the potential of LLMs to discover valid instrumental variables from a large observational database.

\end{abstract}


\input{sections/1_introduction}
\input{sections/2_relatedwork}

\input{sections/3_preliminaries}

\input{sections/4_recoveringIV}

\input{sections/5_IVco-scientist}
\section{Conclusion}
Instrumental variables are central to causal inference in observational studies, but identifying them is challenging and typically demands deep domain expertise. While large language models offer new opportunities for extracting knowledge from text, their use in discovering instruments beyond toy examples remains underexplored. More broadly, we view this setting as part of a new paradigm in which LLMs serve as thought partners for scientific inference, combining their ability to synthesize and reason over unstructured knowledge with the rigor of statistical validation. Within this paradigm, we introduce a multi-agent framework that analyses the data, proposes candidate instruments for a given treatment-outcome pair, and validates them semantically. In addition, we propose a consistency-based metric to assess internal validity in the absence of ground truth. Our empirical results on real-world data demonstrate that LLM-suggested instruments show meaningful consistency, providing a first step toward principled use of LLMs in variable discovery.

\acks{The authors would like to thank Dr. Chuqing Jin and Dr. Alexej Behnisch for their feedback on the instruments proposed by IV Co-Scientist. This work was partially funded by ELSA – European Lighthouse on Secure and Safe AI funded by the European Union under grant agreement No. 101070617. Views and opinions expressed are however those of the authors only and do not necessarily reflect those of the European Union or European Commission. Neither the European Union nor the European Commission can be held responsible for them.}

\section*{Future Work}
A promising direction for future work is to develop more interactive workflows in which LLMs serve as thought partners for causal discovery. Rather than using language models only to propose candidate instruments, future systems could support an iterative process of hypothesis generation, refinement, and validation, combining domain knowledge from LLMs with statistical evidence from data. This opens the door to more collaborative scientific workflows, where LLMs help researchers explore plausible causal structures, suggest new variables, and refine assumptions in tandem with empirical analysis.


\newpage
\bibliography{ref}
\newpage
\appendix
\section{Limitations}
Our approach relies on access to large, structured datasets and uses proxy metrics to evaluate instrument validity, as ground-truth causal effects are rarely observable. While these choices are practical, this could be limiting. Moreover, our framework is tested in relatively well-understood domains and may not generalize to noisier or less-documented fields.

\section{Experimental Setup}
In our paper, we conduct extensive experiments across reasoning and non-reasoning models.  

\subsection{EM and CM}
In Section~\ref{sec:exp1}, we evaluate the ability of LLMs to recover canonical instrumental variables using two metrics, building upon~\cite{sheth2024hypothesizing}: \textbf{Exact Match (EM)} and \textbf{Conceptual Match (CM)}.

\paragraph{Exact Match (EM).}
This metric quantifies the semantic similarity between each LLM-suggested instrument and the known ground-truth instrument. Specifically, we embed both the LLM's suggestion and the literature-sourced IV using the \texttt{Qwen3-Embedding-0.6B} model. We then compute the cosine similarity between the embeddings and report the similarity score, where higher values indicate closer semantic alignment.

\paragraph{Conceptual Match (CM).}
Exact matches may underestimate the utility of LLM suggestions that are plausible but lexically dissimilar. To account for this, we introduce a softer, human-grounded measure: \textbf{Conceptual Match (CM)}. For each LLM-generated IV, we prompt another LLM to act as a domain-aware judge and rate—on a scale from 1 to 10—how conceptually similar the suggestion is to the accepted IV in terms of causal plausibility. A score closer to 10 indicates a stronger conceptual match.

Together, EM and CM allow us to evaluate both surface-level and deeper, contextual alignment between LLM-suggested and literature-backed instruments.

\subsection{Gapminder Dataset}

To evaluate the ability of LLMs to propose and validate novel instrumental variables, we require a rich and diverse source of real-world observational data. For this purpose, we use the Gapminder database\footnote{\url{https://www.gapminder.org/data/}}, a curated compilation of time-series indicators covering over 200 countries and territories.

Gapminder provides over 500 socio-economic and health-related variables, including measures such as GDP per capita, life expectancy, sanitation access, education levels, and fertility rates. These indicators are compiled from authoritative sources like the World Bank, WHO, and UN, and are harmonized to ensure consistency across countries and years. It is under Creative Commons Attribution 4.0.

The diversity and breadth of variables make Gapminder particularly well-suited for causal analysis. It contains plausible treatment and outcome variables across multiple domains (e.g., income, health, demographics), along with a large pool of potential proxy variables. Moreover, the data are longitudinal, enabling time-aware causal reasoning techniques such as Granger causality.

We extract country-year level data for all variables with sufficient temporal coverage. 
To preprocess it, we removed datapoints with missing values and standardized variable scales. This yields a structured dataset suitable for evaluating both traditional statistical tests (e.g., relevance via F-statistic) and novel LLM-generated instruments under realistic conditions.

\subsection{Compute}
We ran Qwen2.5, LLama70b and Llama 8b on A100 GPUs. The GPT models were accessed via API. The PreSelector, HumanProxy and CausalOrale just had to run once, while the discovery modules were iterated over all examples and for new IVs.

\subsection{Reproducibility}
All LLMs were run with a temperature of 0 and top-$p$ of 1 to ensure deterministic outputs. The results reported in Table~\ref{tab:retrival} reflect averaged metrics over multiple runs where applicable. Table~\ref{tab:llm_flawed_iv_evaluation} contains no variance due to LLM randomness, as the models were only used to suggest instruments, which were then evaluated against fixed proxies or ground truth.

We commit to releasing all code, prompts, and evaluation scripts upon acceptance to support full reproducibility.

\section{Qualitative Analysis}
\label{app:qualitative}

To complement our quantitative analyses, we conducted a structured qualitative evaluation with an economics professor and political scientist familiar with instrumental variables. The goal was to assess the plausibility and relevance of LLM-generated variables through expert judgment. The expert was given a short document, consisting of four tasks:

\subsection{Task Overview}

\begin{itemize}
    \item \textbf{Task 1: Generator vs. Critic Evaluation.} The expert was asked whether they agreed with the LLM's rejection of certain candidate instruments following critique by a secondary ``critic'' model that evaluated IVs based on the standard assumptions of relevance, independence, and exclusion.

    \item \textbf{Task 2: Agreement with Accepted and Rejected Instruments.} The expert reviewed a table of treatment–outcome pairs, each with a list of instruments accepted or rejected by the LLM pipeline. They were asked to indicate agreement with each group of variables (e.g., 2/3 accepted IVs, 1/3 rejected IVs).

    \item \textbf{Task 3: Case-Based Evaluation.} The expert was presented with a specific example—female literacy as a treatment for fertility—and asked to comment on the confounders and the plausibility of five candidate instruments, considering both their relevance and threats to validity.

    \item \textbf{Task 4: Reflection on LLMs as Co-Scientists.} The expert reflected on the role of LLMs as collaborators in early-stage IV discovery, and whether such tools might augment, rather than replace, the theoretical reasoning of applied economists.
\end{itemize}

\onecolumn

\section{Detailed Definitions}

\subsection{Granger Causality}
\label{app:granger}
Granger causality is a statistical test used to determine whether one time series is useful in forecasting another. Formally, for time-indexed data $\{v_{i,t}, v_{j,t}\}_{t=1}^T$, we test whether past values of $v_i$ help predict $v_j$ beyond what is possible using past values of $v_j$ alone.

We define the null and alternative hypotheses as follows:

\begin{equation}
    H_0: v_i \text{ does not Granger-cause } v_j
\end{equation}
\begin{equation}
    H_1: v_i \text{ Granger-causes } v_j
\end{equation}

This is operationalized by estimating and comparing the residual variances from two autoregressive models:

\paragraph{Restricted model (without $v_i$):}
\begin{equation}
    v_{j,t} = \alpha_0 + \sum_{k=1}^{p} \alpha_k v_{j,t-k} + \epsilon_t^{(r)}
\end{equation}

\paragraph{Unrestricted model (including lags of $v_i$):}
\begin{equation}
    v_{j,t} = \beta_0 + \sum_{k=1}^{p} \alpha_k v_{j,t-k} + \sum_{k=1}^{p} \gamma_k v_{i,t-k} + \epsilon_t^{(u)}
\end{equation}

The null hypothesis corresponds to testing:
\begin{equation}
    \gamma_1 = \gamma_2 = \dots = \gamma_p = 0
\end{equation}

If the unrestricted model significantly reduces the prediction error compared to the restricted model, we reject $H_0$ and conclude that $v_i$ Granger-causes $v_j$.

\paragraph{Assumptions:}
\begin{itemize}
    \item Both time series are weakly stationary.
    \item The lag length $p$ is appropriately selected.
    \item The model is correctly specified (linearity, no omitted variables).
\end{itemize}

\subsection{ATE Estimation of IVs}

When estimating causal effects using instrumental variables (IVs), we typically recover the \textit{Local Average Treatment Effect} (LATE), not the overall average treatment effect (ATE). This is because IV methods rely on \textit{compliers}—units whose treatment status is affected by the instrument. As a result, the estimated effect pertains only to this subpopulation.

Formally, suppose we have an instrument $Z$, a treatment $T$, and an outcome $Y$. Under the potential outcomes framework, each unit $i$ has:

\begin{itemize}
    \item $T_i(1)$ and $T_i(0)$: potential treatment values if $Z_i = 1$ or $Z_i = 0$
    \item $Y_i(1)$ and $Y_i(0)$: potential outcomes under treatment or no treatment
\end{itemize}

We define the following groups:
\begin{itemize}
    \item \textbf{Compliers:} $T_i(1) = 1$, $T_i(0) = 0$
    \item \textbf{Never-takers:} $T_i(1) = 0$, $T_i(0) = 0$
    \item \textbf{Always-takers:} $T_i(1) = 1$, $T_i(0) = 1$
    \item \textbf{Defiers:} $T_i(1) = 0$, $T_i(0) = 1$ (typically ruled out by the monotonicity assumption)
\end{itemize}

\paragraph{Key Assumptions for LATE:}
\begin{enumerate}
    \item \textbf{Relevance:} $\mathbb{E}[T|Z=1] \ne \mathbb{E}[T|Z=0]$ (instrument affects treatment)
    \item \textbf{Independence:} $Z \perp \!\!\! \perp (Y(0), Y(1), T(0), T(1))$ (instrument is as good as randomly assigned)
    \item \textbf{Exclusion Restriction:} $Z$ affects $Y$ only through $T$ (no direct effect on outcome)
    \item \textbf{Monotonicity:} $T_i(1) \ge T_i(0)$ for all $i$ (no defiers)
\end{enumerate}

Under these assumptions, the LATE is identified as:

\begin{equation}
\text{LATE} = \frac{\mathbb{E}[Y | Z=1] - \mathbb{E}[Y | Z=0]}{\mathbb{E}[T | Z=1] - \mathbb{E}[T | Z=0]}
\end{equation}

This ratio represents the average causal effect of $T$ on $Y$ for compliers only.

\paragraph{Two-Stage Least Squares (2SLS):}  
To estimate LATE in practice, we use a two-stage regression procedure.

\label{appendix:consistency_theory}

In this appendix, we provide additional theoretical insights into the consistency metric introduced in the main text, highlighting its connection to the variance and bias properties of instrumental variable estimators.

\subsection{Consistency as a Measure of Instrument Validity}

Let $Z$ be a candidate instrument used to estimate the causal effect $\beta$ via
\begin{equation}
    \hat{\beta}_{\text{IV}}^{(Z)} = \frac{\text{Cov}(Z, Y)}{\text{Cov}(Z, X)}.
\end{equation}

Assuming $Z$ satisfies the classical instrument validity conditions (relevance and exclusion), the IV estimator is consistent and unbiased in large samples. When multiple valid instruments $Z_1, Z_2, \ldots, Z_m$ are available, their estimates $\hat{\beta}_{\text{IV}}^{(Z_i)}$ should converge to the true causal effect $\beta$ as sample size grows, resulting in low pairwise differences:
\begin{equation}
    \lim_{n \to \infty} \mathbb{E}\left|\hat{\beta}_{\text{IV}}^{(Z_i)} - \hat{\beta}_{\text{IV}}^{(Z_j)}\right| = 0, \quad \forall i,j.
\end{equation}

Large observed discrepancies suggest violations of instrument validity, such as weak instrument bias or direct pathways from $Z_i$ to $Y$ that bypass $X$.

\paragraph{Relation to Instrument Strength and Bias}

The variance of each IV estimate depends inversely on the strength of the instrument, quantified by $\text{Cov}(Z, X)^2$. Weak instruments induce greater variability, leading to increased disagreement between estimates from different instruments.

Additionally, bias from invalid instruments inflates the expected pairwise difference. Formally, for instruments $Z_i$ and $Z_j$, the expected squared difference decomposes as
\begin{equation}
    \mathbb{E}\left[(\hat{\beta}_{\text{IV}}^{(Z_i)} - \hat{\beta}_{\text{IV}}^{(Z_j)})^2\right] = \underbrace{\text{Var}(\hat{\beta}_{\text{IV}}^{(Z_i)}) + \text{Var}(\hat{\beta}_{\text{IV}}^{(Z_j)})}_{\text{variance component}} + \underbrace{(\text{Bias}(Z_i) - \text{Bias}(Z_j))^2}_{\text{bias component}}.
\end{equation}

This decomposition illustrates how the consistency metric reflects both random variation and systematic bias in the set of instruments.

\paragraph{Implications for Instrument Selection}

The normalized consistency score introduced in the main text effectively summarizes these properties by comparing observed discrepancies to a baseline derived from random (invalid) instruments. A low score implies both low variance and low bias among the instruments, supporting their joint validity.

In practice, this metric can guide the selection and refinement of instruments by:
\begin{itemize}
    \item Identifying instruments that cause high disagreement, which may be candidates for exclusion.
    \item Providing a quantitative measure to compare different instrument sets.
    \item Complementing formal tests of instrument validity such as overidentification tests.
\end{itemize}

\section{Rejected IVs}

\begin{table}[ht]
\centering
\begin{tabular}{lll}
\toprule
\textbf{Treatment} & \textbf{Outcome} & \textbf{Rejected IVs} \\
\midrule
GDP & Conflict & Rainfall \\
BMI & SBP & MR  \\
Church attendance & Crime & Rainy days \\
Turnout & Vote share & Rainfall \\
Protests & Prices & Rainfall \\
\bottomrule
\end{tabular}
\label{tab:treatment_outcome_rejected}
\caption{Treatment–Outcome Pairs with Rejected Instruments}

\end{table}

\section{Gapminder preprocessing by IV Co-Scientist}
\label{app:pre}
Table~\ref{tab:preprocessing_summary} presents key preprocessing statistics for each variable pair from Gapminder, including the observed correlation between treatment and outcome variables and the corresponding sample sizes used in the analysis. These metrics provide context on the data quality and strength of associations before causal inference.
\begin{table}[ht]
\centering
\begin{tabular}{llcc}
\toprule
\textbf{Treatment} & \textbf{Outcome} & \textbf{Correlation} & \textbf{Number of Data Points} \\
\midrule
GDP & Health & 0.902 & 2784 \\
Income & Carbon emissions & 0.832 &  1790 \\
Sanitation & Child mortality rate & -0.812 &  2578\\
Poverty & Cholesterol & -0.842 & 3568 \\
Female literacy rate & Number of kids per female & -0.812 &  2504 \\
\bottomrule
\end{tabular}
\caption{Preprocessing Summary: Correlation and Sample Size by Treatment–Outcome Pair}

\label{tab:preprocessing_summary}
\end{table}

\section{IVs generated by IV Co-Scientist}

Table~\ref{tab:ivs_comparison} summarizes the sets of accepted and rejected instrumental variables (IVs) for each treatment–outcome pair, as suggested by GPT-4o. The accepted IVs represent those variables the model deemed more plausible instruments after a critique stage, while the rejected IVs are those filtered out due to likely violations of IV assumptions. 
\begin{table}[ht]
\centering
\scriptsize
\begin{tabular}{llp{4cm}p{5cm}}
\toprule
\textbf{Treatment} & \textbf{Outcome} & \textbf{Accepted IVs} & \textbf{Rejected IVs} \\
\midrule
GDP & Health & 
1. Distance to the port \newline
2. Global commodity prices & 
1. Colonial legal-origin dummies \newline
2. Fertile land \newline
3. Historical settler-mortality rates \\
\midrule
Income & Carbon emissions & 
1. Industrial or resource endowments \newline
2. Policy reforms \newline
3. Trade in country & 
1. Distance to the equator \newline
2. Railroad network density \\
\midrule
Sanitation & Child mortality rate & 
1. Groundwater depth \newline
2. Sewerage investment & 
1. Sanitation subsidy rollout schedule \newline
2. Distance to health center \newline
3. Terrain \\
\midrule
Poverty & Cholesterol & 
1. Cash-transfer age cutoff \newline
2. State minimum wage & 
1. Childcare-program timing \newline
2. State EITC rate \\
\midrule
Female literacy rate & Number of kids per female & 
1. Number of female teachers \newline
2. Raised compulsory school-leaving age \newline
3. Introduction years of a girls-only scholarship program & 
1. Distance to school \newline
2. Historical density of missionary girls’ schools (pre-independence) \newline
3. UI replacement rate \\
\bottomrule
\end{tabular}
\caption{Accepted and Rejected Instruments by Treatment–Outcome Pair}

\label{tab:ivs_comparison}
\end{table}

\newpage
\section{Prompts}
\begin{boxB}{HypothesisGenerator (Instrumental Variable)}
\texttt{
You are an economist helping to identify causal relationships. Given the treatment variable \{T\} and the outcome variable \{Y\}, please provide
a list of 5 possible instrumental variables that could help estimate the causal effect of \{T\} on \{Y\}. The context of this treatment-outcome pair is \{Context\}. These should be variables that influence \{T\} but do not directly affect \{Y\} except through \{T\}. Think step by step. Return    answer with Answer = [list of 5 IVs] }
\end{boxB}

\begin{boxB}{HypothesisGenerator (Confounder)}
\texttt{
You are an economist helping to identify causal relationships. Given the treatment variable \{T\} and the outcome variable \{Y\}, please provide a list of 5 possible confounding variables that might affect both \{T\} and \{Y\}, potentially biasing the causal effect estimate. The context of this treatment-outcome pair is \{Context\}. 
Think step by step. Return your answer with Answer = [list of 5 confounders]}
\end{boxB}

\begin{boxB}{HypothesisGenerator (Independence)}
\texttt{
You are an economist evaluating the validity of instrumental variables. Given the treatment variable \{T\}, outcome variable \{Y\}, a candidate instrumental variable \{Z\}, and a list of confounders \{$U_1, U_2, \ldots$\}, please assess the independence criteria i.e. \{Z\} must be independent of any confounders that affect both \{T\} and \{Y\}. Based on these definitions and the \{Context\}, please evaluate whether \{Z\} is a valid instrument. Think step by step. Return your answer with Answer = [Valid / Invalid]}
\end{boxB}

\begin{boxB}{HypothesisGenerator (Exclusion)}
\texttt{
You are an economist evaluating the validity of instrumental variables. Given the treatment variable \{T\}, outcome variable \{Y\}, a candidate instrumental variable \{Z\}, please assess the exclusion criteria i.e. \{Z\} affects the outcome \{Y\} only through the treatment \{T\}, with no direct effect on \{Y\}. Based on these definitions and the \{Context\}, please evaluate whether \{Z\} is a valid instrument. Think step by step. Return your answer with Answer = [Valid / Invalid].
}
\end{boxB}

\begin{boxB}{ProxyHuman}
\texttt{
You are a policy-minded economist tasked with identifying socio-economically meaningful causal questions. Given a candidate pair \{(T, Y)\}, assess whether: 
1. The relationship is important or interesting i.e., is this a question researchers or policymakers would care about?
2. The pair is interpretable and policy-relevant in real-world socio-economic contexts.
3. The question could plausibly be studied using observational data. 
Avoid pairs that are too similar in meaning (e.g., literacy at ages 5–10 and literacy at ages 10–15). 
Think step by step, using the reasoning a social scientist or economist might apply when deciding whether to pursue this question. 
}
\end{boxB}

\begin{boxB}{CausalOracle}
\texttt{
You are an economist reasoning about causal direction between two socio-economic variables. 
Given a variable pair (A, B) with a strong observed correlation, your task is to determine the likely causal relationship. 
Please evaluate: 
1. Is it more plausible that A causes B? 
2. Is it more plausible that B causes A? 
3. Could the relationship be bidirectional? 
4. Or is the correlation likely driven by confounding or coincidence, with no direct causal link? 
Use real-world knowledge and reasoning as an economist to assess plausibility. Think step by step.
Return your answer as: Answer = [1 / 2 / 3 / 4].
}
\end{boxB}

\end{document}

%% file: sections/1_introduction.tex
\section{Introduction}

%

Understanding the causal effect of a treatment on an outcome is an important question that many disciplines aim to answer~\citep{pearl2009causality}. In economics, we might be interested in understanding: \textit{Does additional schooling increase earnings?}. However, estimating such values is challenging due to endogenity i.e. when the treatment variable is correlated with unobserved factors that also affect the outcome~\citep{bowden1990instrumental, pearl2009causality}. Additionally, often the treatment itself may not be directly measurable or may be subject to measurement error. To address this, instrumental variables (IVs) are used \citep{schennach2016recent, carroll1994measurement,wansbeek2001measurement}. 
A valid IV must causally influence the treatment, must not be an effect of the treatment, and must affect the outcome only through its impact on the treatment. When the conditions for IV are met, IVs allow for consistent estimation of causal effects even in the presence of unobserved confounders. Therefore, identifying or constructing valid instrumental variables is a crucial and often challenging task. The strength and validity of an IV directly determine the reliability of the resulting causal estimates, making this step essential in empirical research for many domains such as economics~\citep{sargan1958estimation, heckman2004using} and health sciences~\citep{cawley2012medical, baiocchi2014instrumental}.

Identifying valid instrumental variables is a highly challenging task that lies at the intersection of theory, domain expertise, and empirical evidence~\citep{jiang2017have}. 
Statistically, a valid instrument must satisfy key conditions: relevance i.e., correlation 
with the endogenous treatment and exclusion i.e. no direct effect on the outcome except through the treatment. Additionally, instruments must be independent of unobserved confounders that affect both treatment and outcome. Satisfying these assumptions often requires more than statistical tests; it demands deep contextual knowledge. Domain experts rely on institutional insights, historical background, policy design features, or mechanisms from natural and social sciences to argue for or against the validity of proposed instruments~\citep{davies2013issues}. This suggests that IV discovery is an inherently multidisciplinary and context-sensitive task. It also requires creativity: plausible sources of exogenous variation are often not directly observable in data and must be carefully hypothesized. Moreover, even instruments once considered valid are sometimes later challenged~\citep{mellon2024rain}. For instance, rainfall has been used as an instrument for studying the effect of war on a country's progress 
but subsequent research has cast doubt on its exclusion validity due to direct effects on the outcome~\citep{sarsons2015rainfall}. Such examples show both the fragility and the difficulty of rigorous instrument identification.

Given these challenges, it is natural to consider whether large language models (LLMs) can assist in IV discovery. Trained on vast textual data spanning domains such as economics, health sciences, law, and history, LLMs have access to a breadth of implicit domain knowledge that could be useful for generating and evaluating candidate instruments~\citep{brown2020language}. Recent work has demonstrated the effectiveness of LLMs in various scientific tasks, including literature reviews~\citep{agarwallitllms, scherbakov2024emergence}, hypothesis generation~\citep{scherbakov2024emergence, agarwallitllms}, and experimental design~\citep{ai4science2023impact, lu2024ai}. This positions LLMs as promising tools to support early-stage IV discovery, not as replacements for theoretical reasoning, 
but as “\textbf{thinking collaborators}” that can augment human intuition and creativity. In this paper, we systematically explore this possibility.

In the context of causal discovery, LLMs have demonstrated strong capabilities in causal discovery, 
often outperforming classical statistical methods~\citep{vashishtha2023causal,kiciman2023causal, anonymous2024causal}. Their contextual reasoning abilities have been leveraged to identify mediators, extract causal graphs from text, and simulate interventions via structured prompting. Building upon these results, we hypothesize that IV discovery is particularly well-suited for LLMs, as it combines both domain-informed causal reasoning and creative hypothesis generation.

In this paper, we investigate whether large language models can assist in the discovery of instrumental variables through a structured, multi-agent framework in which LLM-based agents propose, critique, and refine candidate instruments. Our overarching goal is to evaluate whether LLMs can contribute to the generation of novel instruments for previously unstudied treatment–outcome pairs, rather than merely reproducing known results from the literature. To this end, we ground our evaluation in a real-world, high-dimensional setting using the Gapminder dataset, which contains over 300 socio-economic indicators measured across countries and time.

To build toward this, we adopt a staged evaluation approach. First, we assess whether LLMs can recover well-established instruments from the literature, thereby validating their ability to replicate standard causal reasoning. Second, we examine whether LLMs can correctly avoid suggesting instruments that have been invalidated by prior theoretical or empirical work. Finally, once LLMs demonstrate reliability in these baseline settings, we task them with generating candidate instruments for novel scenarios, offering an opportunity to evaluate their potential in creative and context-sensitive causal discovery. To support empirical evaluation in the absence of ground-truth instrument validity, we additionally introduce a novel metric that measures the internal consistency of causal effect estimates obtained from LLM-suggested instruments. 


%% file: sections/2_relatedwork.tex
\section{Related Work}

\paragraph{LLMs and Causality.}
LLMs have been used as priors to discover the relationships between causal variables~\citep{kiciman2023causal,long2023can,darvariu2024large,ban2023query,vashishtha2023causal, anonymous2024causal}. These methods alone or in combination with statistical or deep learning methods outperformed the latter. 
\citet{sheth2024hypothesizing} proposed a benchmark for discovering causal variables from a partial graph; however, they compared against the ground truth semantically without knowing the statistical effect. ~\cite{verma2025causal} proposed Causal AI Scientist that augmented LLMs with causal tools. The closest work to ours is by ~\citet{han2024mining}, however, our work goes beyond prompting the previously established instruments. 

\paragraph{LLMs and Scientific Discovery.} LLMs are increasingly integrated into various stages of the scientific research workflow, including hypothesis generation and reasoning~\citep{xu2023large, qiu2023phenomenal, si2024can, li2024chain}, coding and implementation~\citep{jimenezswe, barke2023grounded}, data analysis~\citep{ma2023demonstration}, and even peer review~\citep{jin2024agentreview}. ~\citet{chen2026causalevolve} added causal reasoning and causal scratchpad to evolutionary scientific hypothesis search. Despite their growing role, it remains challenging to assess the significance or scientific plausibility of hypotheses generated by LLMs~\citep{lin2025evaluating}, especially when another LLM is used as a judge. In this work, we study LLM-driven discovery of instrumental variables for novel treatment–outcome pairs and propose evaluation metrics to validate their statistical and causal validity.

\paragraph{Instrumental variables.}
Testing an instrumental variable beyond the relevance test is challenging~\citep{sharma2018necessary}, hence we introduce a treatment effect-based consistency metric to quantify the stability of causal estimates across candidate instruments, offering indirect evidence of validity.

%% file: sections/3_preliminaries.tex
\section{Preliminaries}

The instrumental variable enables identification of causal effects in the presence of endogeneity, i.e, when the treatment variable is correlated with unobserved confounders that also influence the outcome. Formally, let $T$ denote the treatment variable, $Y$ the outcome, and $U$ represent unobserved confounders. A valid instrument $Z$ is a variable that influences $T$ but does not directly affect $Y$, except through its effect on $T$.

We consider the structural equations:
\begin{align}
    T &= f(Z, U_T) \\
    Y &= g(T, U_Y)
\end{align}
where $U_T$ and $U_Y$ may be arbitrarily dependent due to shared unobserved variables $U$. In this setup, $Z$ qualifies as an instrumental variable for estimating the causal effect of $T$ on $Y$ if the IV validity conditions~\citep{pearl2009causality} are satisfied.

\subsection{IV Validity Conditions}

\paragraph{Relevance.}
The instrument must be predictive of the treatment. Formally, $Z$ must have a non-zero association with $T$:
\begin{equation}
    \text{Cov}(Z, T) \neq 0
\end{equation}
The relevance condition implies that the function $f(Z, U_T)$
is not a constant in $Z$, hence $Z$ has a causal effect on $T$. In practice, this condition is assessed through the first-stage regression.

\paragraph{Exclusion Restriction.}
The instrument must not affect the outcome through any path other than via the treatment $T$. Formally, this is a \emph{structural} assumption: there is no direct causal path from $Z$ to $Y$ except through $T$. Under the additional assumption that all confounders of $Z$ and $Y$ are observed and included in the covariate set $X$ (i.e., no hidden confounding between $Z$ and $Y$), this implies the conditional independence:
\begin{equation}
Y \perp\!\!\!\perp Z \mid T, X
\end{equation}
In the presence of unobserved confounders, this conditional independence need not hold even if the structural exclusion restriction is satisfied. This assumption is not empirically testable and is typically justified using domain knowledge.

\paragraph{Independence.}
The instrument must be conditionally independent of the unobserved confounders:
\begin{equation}
    Z \perp\!\!\!\perp U \mid X
\end{equation}
This ensures that the variation in $T$ induced by $Z$ is as random concerning the potential outcomes. Similar to the exclusion criteria, independence is also usually argued by domain knowledge.

\subsection{Estimation via Two-Stage Least Squares}

When a valid instrument is available, the causal effect of $T$ on $Y$ can be estimated consistently using Two-Stage Least Squares (2SLS). Following the economics literature \citep{angrist2001instrumental}, we focus on the linear model.
This involves:

\begin{itemize}
    \item \textbf{Stage 1:} Regress $T$ on $Z$ (and covariates $X$) to obtain predicted treatment $\hat{T}$:
    \begin{equation*}
        T = \alpha_0 + \alpha_1 Z + \alpha_2 X + \varepsilon_T
    \end{equation*}
    \item \textbf{Stage 2:} Regress $Y$ on the fitted values $\hat{T}$ (and $X$):
    \begin{equation*}
        Y = \beta_0 + \beta_1 \hat{T} + \beta_2 X + \varepsilon_Y \text{s.t.}\:\:\:\: \varepsilon_Y \perp (Z, X)
    \end{equation*}
\end{itemize}

Under the IV assumptions, the coefficient $\beta_1$ consistently estimates the causal effect of $T$ on $Y$. For the rest of the paper, we assume that the $\varepsilon$ we focus on is linear noise. 

It is important to note that while the relevance condition is statistically testable, the exclusion and independence conditions are not, and must be argued through theory, domain expertise, or natural experiments. This makes the process of identifying valid IVs fundamentally interdisciplinary and often creative. As such, the search for IVs can benefit from tools that integrate reasoning, background knowledge, and flexible hypothesis generation, a role LLMs may be suited to play.

%% file: sections/4_recoveringIV.tex

\section{LLMs for IV Reasoning}
\label{sec:method}
Our goal is to explore whether LLMs can assist in the discovery of novel and valid IVs. We propose a multi-agent pipeline that separates the creative and evaluative stages of IV discovery, mirroring how human researchers hypothesize and then vet candidate IVs.

Given a treatment-outcome pair $(T, Y)$, we define a two-stage LLM-based framework. The first stage is \textit{HypothesisGenerator} where two agents are prompted with a causal query to propose a list of $i$ and $j$ candidate instrumental variables $\{Z_1, \dots, Z_i\}$ and confounders  $\{U_1, \dots, U_j\}$ respectively. 
In addition to hypothesis generation, it is essential to have LLMs as proxy domain experts to argue about statistically untestable conditions. Hence, the second stage includes \textit{CriticAgents}, where two separate LLM agents independently evaluate the validity of the IVs proposed. One agent reasons that the \textbf{exclusion restriction} holds, i.e., whether $Z_i$ affects $Y$ only through $T$. The second agent assesses \textbf{independence}, i.e., whether $Z_i$ is independent of any unobserved confounders $U_j$ 
that influence both $T$ and $Y$.

Each candidate's instrument $Z_i$ receives binary feedback from both the exclusion and independence agents. Only instruments marked as valid by \emph{both} agents are included in the final output:
\begin{equation}
    \mathcal{Z}_{\text{valid}} = \{Z_i \mid \text{Ex}(Z_i) \land \text{Ind}(Z_i)\}
\end{equation}

\begin{table*}[h]
\tiny
\centering
\begin{tabular}{l|cc|cc|cc|cc|cc}
\toprule
\multirow{2}{*}{\textbf{Model}} & \multicolumn{2}{c|}{\textbf{Military service $\rightarrow$ Earning}} & 
\multicolumn{2}{c|}{\textbf{Education $\rightarrow$ Wages}} & 
\multicolumn{2}{c|}{\textbf{Housing $\rightarrow$ Crime}} & 
\multicolumn{2}{c|}{\textbf{Healthcare $\rightarrow$ Mortality}} & 
\multicolumn{2}{c}{\textbf{Migration $\rightarrow$ Wages}} \\
 & EM $\uparrow$
 & CM $\uparrow$
 & EM $\uparrow$
 & CM $\uparrow$
 & EM $\uparrow$
 & CM $\uparrow$
 & EM $\uparrow$
 & CM $\uparrow$
 & EM $\uparrow$
 & CM $\uparrow$
\\
\midrule
GPT-4o        & 0.74 & 1.00 & 0.82 & 1.00 & 0.75 & 0.83 & 0.68 & 0.91 & 0.40 & 0.74 \\
o3-mini       & 0.73 & 1.00 & 0.82 & 1.00 & 0.37 & 0.53 & 0.59 & 0.89 & 0.45 & 0.81 \\
QwQ       & 0.74 & 1.00 & 0.73 & 1.00 & 0.39 & 0.75 & 0.52 & 0.90 & 0.31 & 0.70 \\
Llama3.1 8B   & 0.28 & 0.42 & 0.48 & 0.76 & 0.36 & 0.49 & 0.32 & 0.65 & 0.35 & 0.60 \\
Llama3.1 70B  & 0.61 & 0.84 & 0.67 & 1.00 & 0.59 & 0.75 & 0.52 & 0.83 & 0.57 & 0.77 \\

\bottomrule
\end{tabular}
\caption{Performance of LLMs in recovering canonical instrumental variables across five benchmark treatment-outcome pairs. Exact Match (EM) captures direct or paraphrased mentions of literature-established IVs, while Conceptual Match (CM) identifies plausibly equivalent proxies judged by an LLM critic.}
\label{tab:retrival}
\vspace{-8mm}
\end{table*}

\subsection{Recovering Canonical IVs}
\label{sec:exp1}
It is essential to first assess whether they can recover IVs that are already well-established in the literature before talking about novel instruments. This serves two purposes: (1) it helps calibrate the LLM's alignment with existing scientific knowledge and reasoning, and (2) it provides a baseline for evaluating the model’s ability to reason causally and contextually about treatment-outcome relationships. If an LLM is unable to identify canonical instruments, then relying on it for more speculative and novel discovery becomes difficult to justify.

We curate a benchmark dataset consisting of well-studied treatment-outcome pairs from economics, health sciences, and social sciences, where valid instrumental variables have been previously proposed and accepted in the literature. Each entry in the benchmark includes a treatment variable $T$ (for example, years of schooling), an outcome variable $Y$ (such as future earnings), and one or more canonical instrumental variables $\{Z^{\ast}_1, Z^{\ast}_2, \dots\}$ sourced from peer-reviewed literature.

\begin{table*}[h]
\scriptsize
\centering
\begin{tabular}{l|cc|cc|cc|cc|cc}
\toprule
\multirow{2}{*}{\textbf{Model}} & \multicolumn{2}{c|}{\textbf{GDP $\rightarrow$ Conflict}} & 
\multicolumn{2}{c|}{\textbf{BMI $\rightarrow$ SBP}} & 
\multicolumn{2}{c|}{\textbf{Church $\rightarrow$ Crime}} & 
\multicolumn{2}{c|}{\textbf{Turnout $\rightarrow$ Vote Share}} & 
\multicolumn{2}{c}{\textbf{Protests $\rightarrow$ Prices}} \\
 & HG $\downarrow$
 & Critic $\downarrow$
& HG $\downarrow$
& Critic$\downarrow$
 & HG $\downarrow$
& Critic $\downarrow$
& HG $\downarrow$
& Critic $\downarrow$
& HG$\downarrow$
 & Critic $\downarrow$
\\
\midrule
GPT-4o & 1 & 0 & 1 & 1 & 1 & 0 & 0 & 0 & 1 & 0 \\
o3-mini & 1 & 0 & 1 & 0 & 1 & 1 & 1 & 0 & 1 & 1 \\
QwQ  & 1 & 0 & 1 & 1 & 1 & 0 & 1 & 1 & 1 & 1 \\
Llama3.1 8B      & 0 & 1 & 0 & 1 & 1 & 0 & 0 & 0 & 1 & 0 \\
Llama3.1 70B     & 1 & 0 & 1 & 0 & 1 & 0 & 1 & 1 & 1 & 0 \\

\bottomrule
\end{tabular}
\caption{Performance of different LLMs in identifying flawed instruments across treatment–outcome pairs. HG indicates whether the HypothesisGenerator proposed flawed IV and Critic when CriticAgent successfully picks an invalid IV.}
\label{tab:llm_flawed_iv_evaluation}
\vspace{-5mm}
\end{table*}

\subsection{Avoiding Invalid IVs}
\label{sec:exp2}
While the ability to recover canonical instruments is important, an equally critical aspect of evaluating LLMs for instrumental variable discovery is their sensitivity to invalid instruments. Several variables that were historically proposed as instruments have since been discredited on theoretical grounds or on empirical evidence, typically due to violations of the exclusion restriction or the independence assumption. For example, IVs like rainfall had been used to estimate of the effect of economic activity on civil conﬂict, but later critiques have revealed direct causal paths or unmeasured confounding, undermining their validity~\citep{mellon2024rain}.

In this experiment, we aim to assess whether LLMs can avoid suggesting such invalidated instruments when prompted with the original treatment-outcome pair. This evaluation probes the depth of the model’s reasoning: does it simply retrieve past associations?

We design a multi-stage evaluation framework to assess the robustness of LLMs in handling invalid instruments. Our goal is: (1) to test whether the LLM proposer avoids historically invalid instruments on its own, and (2) to evaluate whether the critic LLM can reliably detect and reject such instruments, even when explicitly introduced.

Given a treatment-outcome pair $(T, Y)$ with a documented invalid instrument $Z^{-}$ (e.g., rainfall), we perform the following steps:
\begin{enumerate}
    \item \textbf{Proposer Behavior.} 
    We prompt the LLM proposer to generate a list of $k$ candidate instruments $\{Z_1, Z_2, \dots, Z_i\}$. We then evaluate whether the model reproduces $Z^{-}$ or semantically equivalent variants.
    This allows us to assess whether the proposer model has internalized the criticisms of certain instruments or simply replicates canonical (yet flawed) examples from the literature.
    
    \item \textbf{Critic Evaluation.} 
    Regardless of Stage 1, we now explicitly inject $Z^{-}$ into the list of candidate instruments. This injected list is:
    \begin{equation}
    \{Z_1, ..., Z^{-} ,...,Z_i\}
    \end{equation}
    We pass this set through the \textit{CriticAgents}, each independently evaluating the instrument on the Exclusion and Independence criteria.
\end{enumerate}

\subsection{Results}
We evaluate a range of benchmark LLMs to assess their ability to propose and critique instrumental variables. For the generation stage, we test both reasoning models: o3-mini~\citep{o3} and QwQ~\citep{hui2024qwen2} and standard models: GPT-4o~\citep{4o}, Llama3.1 8B~\citep{grattafiori2024llama}, and Llama3.1 70B~\citep{grattafiori2024llama}. The \textit{HypothesisGenerator} then evaluates each candidate instrumental variable (IV) from the generated list ${Z_1, \dots, Z_i}$.
along with the \textit{CriticAgent} validating them. The exclusion check is performed independently for each IV, while the independence check is done via comparisons between each IV and the set of hypothesized confounders.
We fix $i=j=5$ as a balance between promoting diversity in generation and maintaining computational efficiency. We prompt all models with an economist persona to elicit appropriate reasoning.

\subsubsection{Recovering Canonical IVs.} 
\label{res:exp1}
We consider treatment, outcome, and instrument tuples $(T, Y, Z^\ast)$ sourced from literature. In particular, the outcome of earnings due to military service~\citep{small2008war}, the effect of education on wages~\citep{imbens2005robust}, housing and its effect on crime~\citep{disney2023does}, healthcare on mortality~\citep{gowrisankaran1999estimating}, and migration's effect on wages~\citep{llull2018effect}.

For each pair $(T, Y)$, we prompt the multi-agent \textit{HypothesisGenerator} and \textit{CriticAgents} to generate a set of LLM validated candidate instruments $\mathcal{Z}_{\text{valid}}$. We then compare these IVs to the known literature instruments $\{Z^{\ast}_1, \dots\}$ using two matching strategies:
\begin{enumerate}
    \item \textbf{Exact Match (EM):} Semantic similarity checks to identify if a known instrument is directly mentioned or closely paraphrased.
    \item \textbf{Conceptual Match (CM):} LLM-judge whether a generated candidate is a plausible conceptual equivalent or proxy to the known instrument.
\end{enumerate}

\autoref{tab:retrival} summarizes the ability of different models to recover well-established instrumental variables across five canonical treatment-outcome settings. We observe that the strongest models: GPT-4o, o3-mini, and QwQ2.5 can recover canonical instruments with high consistency. Across all of the settings, we observe CM rating is higher than EM, because while LLMs often propose valid instruments that align with the underlying causal rationale, they frequently use alternate phrasings or suggest closely related proxies. 

\subsubsection{Avoiding invalid IV}

Given that we have observed positive results in \autoref{res:exp1}, we are interested in evaluating whether LLMs can recognize and avoid historically discredited IVs. We evaluate whether they suggest the negative IV $Z^-$ and whether the \textit{CriticAgents} can filter them out. We filter these $(T, Y)$ from established literature. In particular, the effect of GDP on conflict in a country~\citep{mellon2024rain}, the effect of body mass index on systolic blood pressure~\citep{0497ef202c781093fb10c56d9206d34cc86fb3e5}, the effect of church attendance on crime~\citep{95c6edca216b6a193e47ba73f5c7b99bae4c04ff}, the effect of vote turnout on party vote share~\cite {lal2023much}, and protests on consumer prices~\citep{mellon2024rain}.

In \autoref{tab:llm_flawed_iv_evaluation}, we evaluate how well different models handle flawed instruments. The \textit{HG} column indicates whether the model directly proposed the flawed instrument ($Z^-$), while the \textit{Critic} column captures whether the \textit{CriticAgent} correctly identified and flagged the flaw.

Overall, we see that the \textit{CriticAgent} plays a vital role in safeguarding against invalid instruments. Even when powerful models like GPT-4o and QwQ occasionally suggest flawed variables, the critic is often able to detect and reject them. This highlights the utility of incorporating an automated critic to evaluate statistical validity post hoc. Interestingly, the Llama3.1 8B model appears more conservative, doesn't propose many flawed IVs. However, when such variables are injected, its critic fails to detect the issue.

%% file: sections/5_IVco-scientist.tex
\begin{figure*}[t!]
\centering
    \includegraphics[width=0.8\textwidth]{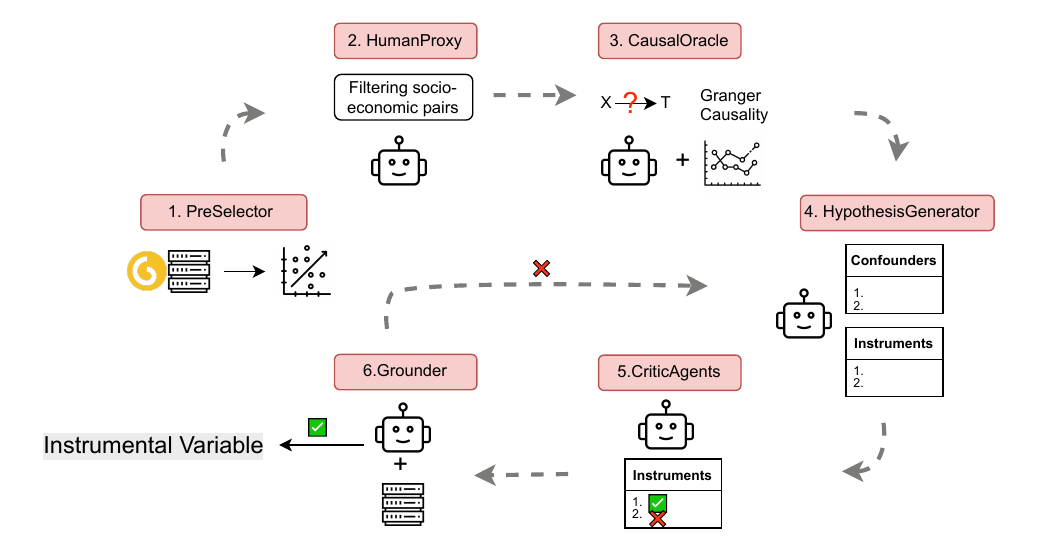}
\caption{Overview of the \textit{IV Co-Scientist} framework, which integrates LLM-based agents with traditional statistical tools.}
    \label{fig:coscientist}
    \vspace{-6mm}
\end{figure*}

\section{IV Co-Scientist}

Having validated the capabilities of LLMs in recovering canonical IVs (\autoref{sec:exp1}) and avoiding discredited ones (\autoref{sec:exp2}), we now evaluate the performance of the system in a fully open-ended setting. The goal here is to test whether LLMs can generate meaningful and potentially novel instrumental variables for real-world causal questions without prior literature as a guide.

This setting reflects a more realistic and challenging scenario: in applied research, analysts often explore large observational datasets to estimate causal effects for which no established IVs exist. In such cases, domain expertise, creativity, and data-driven reasoning are essential to formulating valid IVs. We aim to assess whether LLMs, when paired with a structured evaluation pipeline, can assist in this discovery process.

With this pipeline, we test all of the validity criteria for IV. Similar to the realistic economist pipeline, we evaluate the relevance criteria statistically and the exclusion and independence criteria through LLM's reasoning.

\subsection{Multi-agent IV Discovery}
We use a real-world, high-dimensional sandbox to test open-ended causal exploration. The Gapminder dataset~\citep{rosling2010gapminder} includes socio-economic indicators across countries and over time. The dataset has observations for more than 500 such indicators. We aim to find IVs of novel pairs that are still statistically sound. 

We formulate a multi-stage, multi-agent system, where each agent is responsible for a specific task in the discovery pipeline (see \autoref{fig:coscientist}). Let $\mathcal{V} = \{v_1, v_2, \dots, v_n\}$ denote the set of all variables in the dataset. Our goal is to identify a valid IV $ Z $ for a treatment-outcome pair $ (T, Y) \in \mathcal{V} \times \mathcal{V} $, such that the standard IV conditions are satisfied. Below, we describe the different stages and different agents part of our IV Co-scientist.

\paragraph{Correlation Filter (\textit{PreSelector})}
We compute the Pearson correlation coefficient $\rho(v_i, v_j)$ for all candidate variable pairs, $ (v_i, v_j) \in \mathcal{V} \times \mathcal{V}$. We retain pairs satisfying:
\begin{equation}
\mathcal{P} = \{(v_i, v_j) \mid |\rho(v_i, v_j)| > \tau_\rho, \}
\end{equation}

\noindent for pre-defined thresholds $\tau_{\rho}$. This step eliminates weak or statistically noisy pairs. However, since correlation strength alone does not account for sample size, we also consider the number of data points over which the correlation was computed.

\noindent\textbf{Semantic Relevance Agent (\textit{HumanProxy})}
LLM selects human-meaningful and policy-relevant pairs from $ \mathcal{P} $. Output is set $ \mathcal{S} = \{(v_i, v_j)\}_s $, to be hypothesized as candidate $ (T, Y) $ pairs. This step simulates the reasoning a researcher might apply in choosing interpretable or practically socio-economic questions. Correlation is used here as a necessary but not sufficient filter. 

\noindent\textbf{Causal Direction Agent (\textit{CausalOracle})} 
For each $ (v_i, v_j) \in \mathcal{S} $, apply LLM causal reasoning and statistical tests via Granger causality to infer directionality:
\begin{itemize}
    \item LLM-based Causal Reasoning: Prompted judgments on whether $v_i \rightarrow v_j$ or $v_j \rightarrow v_i$, based on world knowledge.
    \item Granger Causality Test: Statistical test of the temporal data, verifying whether lagged values of $v_i$ improve the prediction of $v_j$, beyond $v_j$’s history. See~\ref{app:granger} for details.
\end{itemize}

We retain only those pairs $(v_i, v_j)$ for which both the LLM and Granger test agree on the direction. The directionally inferred pair is now labeled as $(T, Y)$, with $T$ as treatment and $Y$ as outcome. C

\noindent\textbf{IV Suggestor Agent (\textit{HypothesisGenerator})}
Given a causal pair $(T \rightarrow Y)$, the LLM generates a set of $k$ candidate IVs. See \autoref{sec:method}.

\noindent\textbf{IV Critic Agents (\textit{CriticAgents}) }
Each candidate $Z_i$ is passed through two critic agents that critique the IVs and give a list of $\mathcal{Z}_{\text{valid}}$. See \autoref{sec:method}.

\noindent\textbf{Proxy Matching Agent (\textit{Grounder})}

For each valid IV $Z_i \in \mathcal{Z}_{\text{valid}}$ proposed by the LLM, we attempt to ground it in the dataset by identifying a concrete proxy variable. 
If no such proxy is found, $Z_i$ is excluded from downstream evaluation. Otherwise, the discovered IV is retained as $(Z_i, \text{Proxy}(Z_i))$.

The IV Co-scientist operates as a sequential, dependency-aware pipeline in which each agent refines the output of the previous stage. The process begins with the \textit{PreSelector}, which filters variable pairs using correlation strength of $0.7$ following ~\citet{akoglu2018user} and effective sample size, producing a tractable set of statistically plausible associations. The \textit{HumanProxy} then narrows this set by selecting socio-economically meaningful and policy-relevant pairs, yielding a collection of candidate causal relationships $\mathcal{S}$. Next, the \textit{CausalOracle} combines LLM-based causal reasoning with Granger causality tests to infer directionality; only pairs for which both components agree are retained and labeled as treatment–outcome pairs $(T \rightarrow Y)$.

Given each directed causal pair, the \textit{HypothesisGenerator} proposes a set of candidate instrumental variables intended to isolate exogenous variation in $T$. These proposals are subsequently evaluated by the \textit{CriticAgents}, which assess instrument plausibility with respect to relevance, exclusion, and independence, producing a filtered set of candidate instruments. Finally, the \textit{Grounder} attempts to map each surviving candidate instrument to a concrete proxy variable in the dataset. Only instruments that can be grounded in observed data are passed to downstream causal estimation. If no grounded instruments remain for a given $(T, Y)$ pair, the IV generation and evaluation stages are re-invoked, allowing the system to iteratively explore alternative hypotheses.

\subsection{Evaluation}

Given that the discovered $(T, Y, Z)$ triplets in our open-ended pipeline are novel, direct comparison to ground truth IVs is not feasible. To evaluate the plausibility and effectiveness of the LLM-suggested IVs, we use the statistical strength of the IV, a standard measure that is used in the IV literature. Further, we propose a novel metric to compare sets of valid and invalid IVs. 

\subsubsection{Statistical Strength via F-statistic}

A key requirement for a valid IV is \textit{relevance}, which means that the IV must be sufficiently correlated with the treatment variable. To quantify this, we compute the first-stage F-statistic, a standard method used in instrumental variables analysis to detect weak IVs. Specifically, we regress the treatment variable $T$ on the candidate IV $Z$ and assess whether $Z$ explains significant variation in $T$. A high F-statistic indicates strong predictive power. 

In our analysis, we use robust heteroskedasticity-consistent estimators that do not assume Gaussian errors, reflecting the potentially complex and noisy nature of observational data.
The F-statistic tests the null hypothesis $H_0: \beta = 0$. An F-statistic value above the conventional threshold (typically 10) indicates a strong IV.


\subsubsection{Consistency of Estimated Effects}

While the relevance of an instrument can be directly assessed via predictive strength (e.g., F-statistic), its overall validity also depends on the more elusive \textit{exclusion} and \textit{independence} assumptions. In the absence of ground-truth causal effects, verifying these assumptions directly is not possible. To address this, we propose a novel evaluation metric, which we call \textbf{consistency}, aimed at comparing the quality of IVs sets.

The key idea is as follows: if a set of instruments truly isolates exogenous variation in the treatment, then each instrument in the set should produce similar estimates of the average treatment effect (ATE) when used in a two-stage least squares (2SLS) estimation. A related intuition appears in the econometric literature on overidentification tests, which use multiple instruments to assess whether the implied identifying restrictions are mutually compatible \citep{hausman1978specification,sargan1958estimation}.  In other words, their causal estimates should be \textit{internally consistent}. 

Formally, for two instruments $Z_1$ and $Z_2$ proposed by the LLM, we compute their respective 2SLS estimates of the ATE, and define the consistency score:

\begin{equation}
    \Delta_{\text{LLM}} = \left| \hat{\beta}_{\text{ATE}}^{(Z_1)} - \hat{\beta}_{\text{ATE}}^{(Z_2)} \right|
\end{equation}

A smaller $\Delta_{\text{LLM}}$ indicates greater agreement between the causal estimates and, by extension, stronger internal validity of the instrument set.
To contextualize this metric, we construct a null distribution by randomly sampling proxy variables $R_1$ and $R_2$ from the dataset variables. These random proxies serve as a negative control, capturing the expected variability in treatment effect estimates from invalid or spurious instruments. Our goal is to test whether the consistency observed in the LLM-suggested instruments is significantly better than that from random variables: i.e., whether $\Delta_{\text{LLM}} < \Delta_{\text{Rand}}$.

\begin{equation}
    \Delta_{\text{Rand}} = \left| \hat{\beta}_{\text{ATE}}^{(R_1)} - \hat{\beta}_{\text{ATE}}^{(R_2)} \right|
\end{equation}

This approach is inspired by the self-compatibility test introduced in causal discovery~\citep{faller2024self}, for evaluation in the absence of ground truth. While it does not confirm the validity of the instrument, it provides indirect evidence of the quality. Consistent estimates across diverse LLM-suggested IVs suggest that they may be isolating exogenous variation in the treatment.

This test examines whether the causal effect estimates generated by LLM-suggested IV exhibit greater internal consistency than would be expected by chance. If so, it provides indirect evidence that the LLM is identifying variables that capture meaningful exogenous variation, even when the exclusion and independence assumptions cannot be directly verified.


\subsection{Results}
We perform experiments on the Gapminder database that contains many socio-economic variables, totaling over 500 variables. In addition to standard socio-economic measures, it includes policy-related variables (e.g., CPIA governance indicators, debt and fiscal policy metrics, and over ten government financial indicators) as well as health and biomedical variables, such as cancer, vaccine coverage, malaria, and maternal health outcomes. This breadth ensures that our framework evaluates LLMs across a diverse set of variables and causal contexts. 
\begin{table*}[]
\tiny
    \centering
    \begin{tabular}{l|cc|cc|cc|cc|cc}
    \toprule
        & \multicolumn{2}{c|}{\textbf{GDP $\to$ Health}} 
        & \multicolumn{2}{c|}{\textbf{Income $\to$ Emissions}} 
        & \multicolumn{2}{c|}{\textbf{Sanitation $\to$ Mortality}} 
        & \multicolumn{2}{c|}{\textbf{Poverty $\to$ Cholesterol}} 
        & \multicolumn{2}{c}{\textbf{Female literacy $\to$ Kids}}\\
        & Relevance & $\mathcal{C}_{\text{norm}}$  
        & Relevance & $\mathcal{C}_{\text{norm}}$  
        & Relevance & $\mathcal{C}_{\text{norm}}$  
        & Relevance & $\mathcal{C}_{\text{norm}}$  
        & Relevance & $\mathcal{C}_{\text{norm}}$ \\
    \cmidrule{1-3} \cmidrule{4-5} \cmidrule{6-7} \cmidrule{8-9} \cmidrule{10-11}

GPT-4o     & 14.28 & 0.515 & 17.52 & 0.524 & 11.37 & 0.508 & 13.44 & 0.532 & 19.81 & 0.518 \\
o3-mini     & 14.28 & 0.515 & 14.88 & 0.505 & 11.37 & 0.508 & 10.32 & 0.568 & 19.81 & 0.518 \\
QwQ         & 13.10 & 0.541 & 16.05 & 0.529 & 10.76 & 0.515 & 12.51 & 0.541 & 18.23 & 0.529 \\
Llama3.1 70b& 12.65 & 0.513 & 15.33 & 0.529 & 10.28 & 0.581 & 11.90 & 0.538 & 17.50 & 0.559 \\
Llama3.1 8b & 13.10 & 0.541 & 11.92 & 0.559 & 10.76 & 0.515 & 12.51 & 0.541 & 18.23 & 0.529 \\

    \midrule
    \end{tabular}
    \caption{Performance of different LLMs in discovering novel IVs. Relevance is defined by the first stage F-statistic and $\mathcal{C}_{\text{norm}}$ gives the consistency when compared to random IVs.}
    \label{tab:novel}
    \vspace{-7mm}
\end{table*}

In \autoref{tab:novel}, we evaluate the quality of LLM-suggested IVs using two complementary perspectives: statistical relevance and consistency of estimated causal effects. An IV is considered strong if it predicts variation in the treatment variable $T$, with an $F$-statistic exceeding a conventional threshold (typically $F > 10$). We consider five examples that were autonomously generated using our multi-agent IV co-scientist. Please refer to the Appendix \ref{app:pre} for further details.

We define a normalized consistency score $\mathcal{C}_{\text{norm}} = |\frac{\Delta_{\text{LLM}}}{\Delta_{\text{Rand}}}|$ which quantifies the stability of causal estimates produced by LLM-suggested IVs relative to random proxies. Values $ < 1$ would suggest that LLM IVs are more consistent and potentially more robust than random variables. 
\begin{figure}[t]
\centering
    \begin{subfigure}
         \centering
        \includegraphics[width=0.3\linewidth]{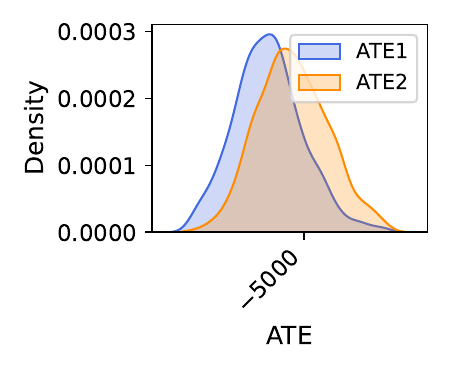}
         \vspace{-1mm}
     \end{subfigure}
     \begin{subfigure}
         \centering
         \includegraphics[width=0.3\linewidth]{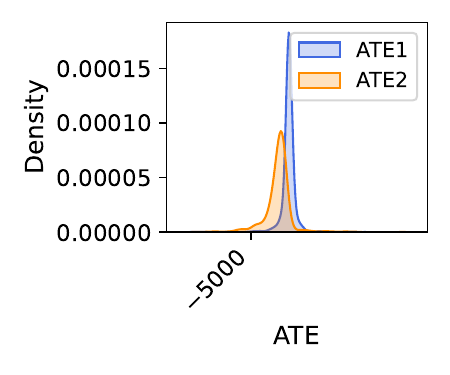} 
         \vspace{-1mm}
     \end{subfigure}
\caption{Comparison of the ATE density while using two different IVs: (a) LLM proposed and (b) random. This is for Sanitation $\rightarrow$ Mortality for GPT-4o.}
\label{fig:ate_posteriors}
\vspace{-6mm}
\end{figure} 
We observe that many of the IVs suggested by the LLM have a high relevance score. We note that while we might observe a statistically significant relationship, this doesn't guarantee the relationship is strong enough to avoid problems with weak IVs~\citep{lewis2022robust}. Therefore, we also focus on the introduced consistency $\mathcal{C}_{\text{norm}}$ score. Empirically, we find that $\mathcal{C}_{\text{norm}}$ are less than $1$, demonstrating a clear gap between the performance of LLM-suggested and random IVs. We note that since, GPT-4o and o3-mini had similar IV suggestions, it led to the same results. 
\autoref{fig:ate_posteriors} visually supports this interpretation. In (a), using LLM-suggested IVs, the posterior distributions for ATE1 and ATE2 suggest both IV relevance and the presence of meaningful local treatment effect heterogeneity. In contrast, panel (b) implies that the IVs may be weak or invalid and highly inconsistent. 

\subsubsection{Ablation}

\paragraph{Sensitivity to the number of proposals.}
We analyze the sensitivity of IV Co-scientist to the number of candidate instruments $i$ and confounders $j$ used to assess independence.
\begin{table*}[h]
\scriptsize
\centering
\begin{tabular}{cc|cc|l}
\toprule
\# Instruments (\(i\)) & \# Confounders (\(j\)) & Relevance (F) & \(\mathcal{C}_{\text{norm}}\) & Notes \\
\midrule
5  & 1  & 8.27  & 0.62 & Independence criteria insufficiently evaluated \\
5  & 5  & 14.28 & 0.52 & Default setting; balanced performance \\
5  & 10 & 13.57 & 0.51 & Independence checks overly restrictive \\
10 & 5  & 16.49 & 0.48 & Highest relevance, increased token cost \\
10 & 10 & 13.57 & 0.51 & No clear improvement over default \\
\bottomrule
\end{tabular}
\caption{Ablation study on the number of candidate instruments (\(i\)) and confounders (\(j\)) in the IV Co-scientist framework using GPT-4o on the GDP \(\rightarrow\) Health causal pair.}
\label{tab:ablation_ij}
    \vspace{-3mm}
\end{table*}

In \autoref{tab:ablation_ij}, we observe that when the number of confounders is small (\(j = 1\)), critic agents lack sufficient context to meaningfully evaluate the independence criterion, resulting in weaker relevance and higher variability in estimated effects. Increasing the number of confounders beyond the default (\(j = 10\)) makes independence checks overly conservative, suppressing plausible instruments without improving consistency. Increasing the number of proposed instruments (\(i = 10\)) can improve relevance but substantially increases token usage with only marginal gains in consistency. Overall, the default configuration (\(i = 5, j = 5\)) provides a favorable trade-off between computational efficiency and instrument quality, with larger settings yielding diminishing returns or stricter filtering without clear downstream benefits.

\paragraph{Interpreting the Consistency Score.} To better understand the meaning of the consistency score, we compare it against a task-specific null distribution constructed by sampling 2000 random proxy pairs from the same dataset and recomputing the statistic. This provides a baseline for the level of agreement expected under random selection. We then evaluate where the LLM-proposed instruments fall within this distribution using a one-sided test. In \autoref{tab:consistency_null}, across tasks, the proposed instruments tend to lie toward the lower tail of the null, indicating stronger agreement than would typically arise from random variable pairs. 
\begin{table}[t]
\centering
\begin{tabular}{l|cc}
\toprule
Task & Consistency & P-value \\
\midrule
GDP $\rightarrow$ Health & 0.515 & 0.072 \\
Income $\rightarrow$ Emissions & 0.524 & 0.048 \\
Sanitation $\rightarrow$ Mortality & 0.508 & 0.033 \\
Poverty $\rightarrow$ Cholesterol & 0.532 & 0.051 \\
Female literacy $\rightarrow$ Kids & 0.518 & 0.061 \\
\bottomrule
\end{tabular}
\caption{Consistency scores evaluated against task-specific null distributions. P-values correspond to tests against random proxy pairs.}
\vspace{-5mm}
\label{tab:consistency_null}
\end{table}

\paragraph{Human evaluation.} We consulted a faculty-level economist to qualitatively assess the LLM-generated IVs (see Appendix \ref{app:qualitative}). They found the \textit{CriticAgents}’ reasoning and confounder identification generally sound. They noted that accepted and rejected IVs often differ not in validity but in generality: accepted ones tend to be broader and less debated, while rejected ones are more specific and often come with known critiques.
